\documentclass{article}

\pdfoutput=1

\usepackage{microtype}
\usepackage{graphicx}
\usepackage{booktabs} %

\usepackage{hyperref}
\usepackage{amsfonts}       %

\usepackage{amsmath}
\usepackage{amsthm}
\usepackage{xcolor}
\usepackage{multirow}

\usepackage{subcaption}
\newcommand{\Tau}{\mathcal{T}}

\usepackage[accepted]{icml2021}

\icmltitlerunning{Regularization Strategies for Quantile Regression}

\begin{document}

\twocolumn[

\icmltitle{Regularization Strategies for Quantile Regression}

\icmlsetsymbol{equal}{*}

\begin{icmlauthorlist}
\icmlauthor{Taman Narayan}{goo}
\icmlauthor{Serena Wang}{goo}
\icmlauthor{Kevin Canini}{goo}
\icmlauthor{Maya R. Gupta}{goo}
\end{icmlauthorlist}

\icmlaffiliation{goo}{Google Research, Mountain View, California, USA}

\icmlcorrespondingauthor{Taman Narayan}{tamann@google.com}
\icmlcorrespondingauthor{Serena Wang}{serenawang@google.com}

\vskip 0.3in
]

\printAffiliationsAndNotice{}  %

\begin{abstract}
We investigate different methods for regularizing quantile regression when predicting either a subset of quantiles or the full inverse CDF. We show that minimizing an expected pinball loss over a continuous distribution of quantiles is a good regularizer even when only predicting a specific quantile. For predicting multiple quantiles, we propose achieving the classic goal of non-crossing quantiles by using deep lattice networks that treat the quantile as a monotonic input feature, and we discuss why monotonicity on other features is an apt regularizer for quantile regression. We show that lattice models enable regularizing the predicted distribution to a location-scale family.  Lastly, we propose applying rate constraints to improve the calibration of the quantile predictions on specific subsets of interest and improve fairness metrics.  We demonstrate our contributions on simulations, benchmark datasets, and real quantile regression problems.
\end{abstract}

\section{Introduction}
Real world users often seek estimates of the quantiles of a random variable. For example, a delivery service may want the estimate that 90$\%$ of the time ($\tau = 0.9$), a delivery will arrive within 38 minutes. For a random variable $Y \in \mathbf{R}$ and a desired quantile $\tau \in (0,1)$, the $\tau$-quantile of $Y$ is defined as $q_\tau = \inf \{q : P(Y \leq q) \geq \tau\}$. For a pair of random variables $(X, Y) \in \mathbf{R}^D \times \mathbf{R}$, the \emph{conditional} $\tau$-quantile of $Y$ for feature vector $X$ is defined as $q_\tau(x) = \inf \{q : P(Y \leq q | X = x) \geq \tau\}$. For example, that conditioned on it being 5pm, the model estimates that $90\%$ of the time it will arrive in 53 minutes.  Quantile regression takes a set of pairs $(x,y) \in \mathbf{R}^D \times \mathbf{R}$ from a joint distribution over $(X, Y)$ as training data and seeks to estimate some or all of the conditional quantiles of $Y$ for any value of $X$. 

Estimating multiple quantiles $\tau$ can produce the classic problem of \emph{quantile crossing} \cite{KoenkerBassett:78}, where the estimated quantiles violate the commonsense requirement that $q_\tau(x) \geq q_{\tau'}(x)$ for $\tau \geq \tau'$ at every $x$.  For example, if an AI tells a  customer there is a $90\%$ chance their delivery will arrive within 38 minutes, but also that there is an $80\%$ chance it will arrive within 40 minutes, then the customer may think the AI is broken or buggy. In fact, this kind of embarrassing mistake happens easily \cite{He:1997,KoenkerBassett:78}; see Table \ref{tab:sim_dist} for data. In this paper, we simultaneously estimate the desired quantiles (or all quantiles) with explicit monotonicity constraints on $\tau$ to guarantee non-crossing quantiles, and explore other  regularization strategies for quantile regression. 

Our main contributions are: (1) showing that training with an \emph{expected} pinball loss can regularize individual quantile predictions, (2) proposing training arbitrarily flexible models that guarantee the classic interpretability goal of \emph{non-crossing quantiles} by treating the quantile as a monotonic feature in a \emph{deep lattice network} (DLN), (3) showing the importance of monotonic regularizers on input features for quantile regression, (4) showing the DLN function class enables regularization of the estimated distribution to a location-scale family, and (5) showing that \emph{rate constraints} can be applied to promote well-calibrated quantile estimates for relevant subsets of the data and to improve fairness metrics. We use a broad set of simulated and real data to illustrate the significance of these regularizers.

\section{Quantile Regression Training Objective} \label{sec:bestever}
First, we give a comprehensive training objective for quantile regression that brings together and modernizes many great insights from the machine learning and statistics communities. Details of the different aspects of this optimization problem, our proposals, and the related work will follow. 

Given a training set $\{(x_i, y_i)\}_{i=1}^N$ where each $x_i \in \mathbf{R}^D$ is a feature vector and $y_i \in \mathbf{R}$ is a corresponding label, we add the quantile of interest $\tau \in (0,1)$ as an auxiliary feature and model the conditional quantile $q_\tau(x)$ by $f(x, \tau; \theta)$, where our model $f:\mathbf{R}^{D+1} \rightarrow \mathbf{R}$ is parameterized by $\theta \in \mathbf{R}^p$. To fit $f$, we propose minimizing the \emph{pinball loss} \cite{KoenkerBassett:78}, $L_\tau(y, \hat{y}) = \max(\tau (y - \hat{y}), (\tau-1)(y - \hat{y}))$, in expectation with respect to a distribution $P_\Tau$, where $\Tau \in (0, 1)$ is a random quantile, subject to constraints that quantile estimates may not cross. We also propose optionally adding \emph{rate constraints} \cite{Goh:2016,CotterJMLR} to ensure the quantile estimates are well-calibrated on specified data subsets. The resulting problem is a constrained optimization:
\begin{align}
&\min_{\theta} \mathbb{E}_\Tau \left[ \sum_{i=1}^N L_\Tau(y_i, f(x_i, \Tau; \theta)) \right] \label{eqn:objective}\\
& \textrm{s.t. } f(x, \tau^+; \theta) \geq f(x, \tau^-; \theta) \label{eqn:peace} \\ 
& \forall x \in \mathbf{R}^D \textrm{ and any } \tau^+, \tau^- \in (0, 1) \textrm{ where } \tau^+ \geq \tau^-, \nonumber \\
& \textrm{s.t. } (\tau_s - \epsilon^-_s) \leq  \frac{1}{|\mathcal{D}_s|}\sum_{(x_j, y_j) \in \mathcal{D}_s}\mathbb{I}[y_j \leq f(x_j, \tau_s; \theta)] \nonumber \\
& \hspace{20mm} \leq (\tau_s + \epsilon^+_s) \textrm{ for } s \in \{1, \ldots, S\}, \label{eqn:fairness}
\end{align}
Each of the $S$ \emph{rate constraints} in (\ref{eqn:fairness}) are specified by a quantile $\tau_s$, a dataset $\mathcal{D}_s$ of interest, which may be a subset of the training data or an auxiliary dataset, and allowed slacks $\epsilon^-_s, \epsilon^+_s \in [0, 1]$. $\mathbb{I}$ is the usual indicator.

\section{Minimizing the Expected Pinball Loss}
\label{sec:expected-pinball}
The idea in (\ref{eqn:objective}) and (\ref{eqn:peace}) of simultaneous estimation of quantiles by minimizing the sum of pinball losses over a \emph{pre-specified, discrete set} of quantiles with non-crossing constraints dates to at least \citet{Takeuchi:2006} and is generally regarded as useful \cite{Bondell:2010,LiuWu:2011,Bang:2016,Cannon:2018}. \citet{SQR:2019} proposed training for \emph{every} $\tau$ equally by minimizing the \emph{expected} loss over \emph{all} quantiles, drawing a separate $\tau$ uniformly at random between zero and one for each example in each batch or epoch of training. \citet{SQR:2019} show that minimizing with a uniform $P_\Tau$ expected pinball loss induces smoothing across $\tau$ that likely \emph{reduces} non-crossing, but we show in Section~\ref{sec:experiments} that crossing can still occur quite frequently with DNNs.

We note that these prior works, and the case of single quantile regression, can all be understood as special cases of choosing a distribution $P_\Tau$ to sample the pinball loss from and then optimizing (\ref{eqn:objective}). In fact, we will show that using a broad $P_\Tau$ regularizes the estimates of specific $\tau$'s, particularly closer to the median.  Thus, even if only a single $\tau$ is of interest, we propose using a beta distribution for $P_\Tau$ centered on that $\tau$ as a regularization strategy. Likewise, if one is only interested in a few discrete quantiles, we will show that training with a \emph{uniform} $P_\Tau$ can be a good regularizer. 

This proposal extends what statisticians have long known to be true for unconditional quantile estimation: that quantile estimators that smooth multiple quantiles can be more efficient than simply taking the desired sample quantile \cite{HarrellDavis:1982,Kaigh:1982,David:1986}. For example, for a uniform $[a,b]$ distribution, the average of the sample min and sample max is the minimum variance unbiased estimator (MVUE) for the sample median (follows from the Lehmann-Schaffe theorem).  The popular Harrell-Davis quantile estimator is a weighted average of all the sample order statistics, and is asymptotically equivalent to computing the mean of bootstrapped medians \cite{HarrellDavis:1982}.

\section{Why Use Deep Lattice Networks?}
We propose using deep lattice networks (DLNs) \cite{You:2017} for quantile regression, because they can be made arbitrarily flexible, and as we show in this section, DLNs efficiently enable three key regularization strategies: non-crossing constraints, monotonic features, and restricting the learned distribution to location-scale families. 

\subsection{Lattice Models}
A lattice is a nonlinear function formed by interpolating a multi-dimensional look-up table \citep{Garcia:09}. Lattices are naturally smooth and can approximate any continuous bounded function by adding more keypoints to the multi-dimensional look-up table. Because the look-up table parameters form a regular grid of function values, many shape constraints such as monotonicity can be imposed efficiently through linear inequality constraints on the parameters\cite{GuptaEtAl:2016,canini:2016,You:2017,Gupta:2018,SetConstraintsICML2019}. DLNs are state-of-the-art universal approximators for bounded partially monotonic functions \cite{You:2017}, and are made up of lattice layers, linear layers, and calibrator layers (which are a set of one-dimensional piecewise linear functions). See the Supplemental for a quick review on DLNs.

\subsection{Achieving Non-crossing Quantiles}
The non-crossing constraints of (\ref{eqn:peace}) encode a common-sense expectation that aides model interpretability and trustworthiness, as well as serving as as a semantically-meaningful, tuning-free regularizer. We propose using DLNs to achieve non-crossing quantiles. First, add $\tau$ as an input feature to the DLN as done in (\ref{eqn:objective}), and thus the resulting quantile regression function $f(x,\tau)$ can represent arbitrary bounded quantile functions, given enough model parameters.  Second, impose monotonicity on the $\tau$ parameter in the DLN to satisfy (\ref{eqn:peace}), while still achieving full flexibility for other features. Our implementation uses the open-source TensorFlow Lattice library~\cite{TFLatticeBlogPost2020}. More generally, one could use unrestricted ReLU or embedding layers for the first few model layers on $x$, then fuse in $\tau$ later with $\tau$-monotonic DLN layers. 

Prior work has used a similar mechanism to impose the non-crossing constraint (\ref{eqn:peace}) for a \emph{discrete} number of quantiles and for \emph{more-restrictive} function classes, such as linear models \cite{Bondell:2010} and two-layer monotonic neural networks \cite{Cannon:2018}, which are known to have limited flexibility \cite{Minin:2010,Daniels:2010}. Models with more layers \cite{Minin:2010,Lang:2005} or min-max layers \cite{Sill:98} can provide universal approximations. Monotonic neural nets have also been proposed for estimating a CDF \cite{Silva:2018}.  

\subsection{Monotonic Features for Quantile Regression}
DLNs can efficiently impose monotonicity on selected input features. Monotonicity constraints are particularly useful for quantile regression because real-world quantile regression problems often use features that are past measurements (or strong correlates) to predict the future distribution of measurements. For example, when predicting the quantiles of the time a bus route takes given $D=8$ features, its past 7 travel times and the month, if any of the past travel time features were increased, the model should predict longer future travel times, never shorter ones. This type of domain knowledge can be captured as a tuning-free, semantically-meaningful regularizer that also aids model interpretability by constraining the model's predictions to be monotonically increasing in each of those input features. 

\subsection{Regularizing to Location-Scale Distributions}
Using a DLN for $f$, and training (\ref{eqn:objective}) with a uniform $P_{\Tau}$ expected pinball loss and imposing non-crossing constraints as per (\ref{eqn:peace}), our method will estimate a complete and well-behaved inverse CDF, %
unlike much of the prior empirical risk minimization work in quantile regression. \cite{Cannon:2018,SQR:2019,Bondell:2010,Takeuchi:2006,LiuWu:2011,Bang:2016}. 

There are two main prior approaches to estimating a complete inverse CDF. The first relies on nonparametric strategies; K-nearest neighbor methods, for example, can be extended to predict quantiles by taking the quantiles rather than the mean from within a neighborhood \cite{QKNN:1990}. Quantile regression forests \cite{Meinshausen:2006} use co-location in random-forest leaf nodes to generate a local distribution estimate. 

The second strategy is to fit a parametric distribution to the data. Traditionally these methods have been fairly rigid, such as assuming Gaussian noise. \citet{He:1997} developed a method to fit a shared but learned location-scale family across $x$. \citet{Yan:2018} found success with a modified 4-parameter Gaussian whose skew and variance was dependent on $x$. Recently, \citet{Gasthaus} proposed \emph{spline quantile function DNN models} whose outputs are the parameters for a piecewise-linear quantile function, which can fit any continuous bounded distribution, given sufficient parameters. They only discuss recurrent neural networks, but their framework is applicable to the generic quantile regression setting we treat as well. 

One advantage of our approach, with DLNs and an explicit $\tau$ feature, is that it maintains the possibility of full flexibility, as do the nonparametric methods and \citet{Gasthaus}, but enables regularizing the distribution in very natural ways by making certain architecture choices, similar to the more rigid distributional approaches. As a simple example, by constraining the DLN architecture to not learn interactions between $\tau$ and any of the $x$ features, one learns a regression model with homoskedastic errors. In fact, a basic two-layer DLN called a \emph{calibrated lattice} model \cite{GuptaEtAl:2016} can be constrained to learn distributions across $x$ that come from a shared, learned, location-scale family:

\textbf{Lemma:} Let $f(x, \tau)$ be a calibrated lattice model \cite{GuptaEtAl:2016} with piece-wise linear calibrator $c(\tau): [0,1] \rightarrow [0,1]$ for $\tau$, and only 2 look-up table parameters for $\tau$ in the lattice layer, and suppose the look-up table is interpolated with multilinear interpolation to form the lattice. Then $f(x,\tau)$ represents an inverse CDF function $F^{-1}(y|x)$ where the estimated distribution for every $x$ is from the same location-scale family as the calibrator $c(\tau)$.

\textbf{Proof:} If a random variable $Y$ conditioned on $X$ belongs to the location-scale family then for $\tau \in (0,1)$ and $a \in \mathbb{R}$ and $b>0$, it must hold that the conditional inverse CDF satisfies $F^{-1}_{Y|X=z}(\tau) = a + b F^{-1}_{Y|X=x}(\tau)$.  Note that for  $\tau \in (0,1)$, interpolating a lattice with two look-up table parameters in the $\tau$ dimension yields the estimate $\hat{F}^{-1}_{Y|X=z}(\tau) = f(z,\tau) = f(z, 0) + c(\tau) (f(z, 1) - f(z, 0))$. Thus mapping to the location-scale property, $a = f(z, 0)$,  $b = f(z, 1) - f(z, 0)$, and $\hat{F}^{-1}_{Y|X=x}(\tau) = c(\tau)$. Thus every estimated conditional inverse CDF  $\hat{F}^{-1}_{Y|X=z}(\tau)$ is a translation and scaling of the piecewise linear function $c(\tau)$.

The number of keypoints in $c(\tau)$ controls the complexity of the learned base distribution, allowing the model to approximate location-scale families like the Gaussian, gamma, or Pareto distributions. The number of lattice knots in the $\tau$ feature, meanwhile, naturally controls how much the distribution should be allowed to vary across $x$. Two knots, as noted above, limits us to a shared location-scale family, while three knots gives an extra degree of freedom to shrink or stretch one side of the distribution differently across $x$. Ensembling, layer depth, and further lattice vertices in $\tau$ steadily move one towards full generality.

\section{Rate Constraints and Quantile Property}
\label{sec:rate_constraints}
Quantile regression models would ideally satisfy the \emph{quantile property} \cite{Takeuchi:2006}, meaning that the proportion of observed outcomes less than the model prediction is $\tau$, for \emph{any subset} of the data.  This subset accuracy issue also a primary concern for fairness in machine learning: for example, one may wish to ensure quantile estimates achieve some mandated level of quantile accuracy for each of a set of socioeconomic groups.

Prior work \cite{Takeuchi:2006,Sangnier:2016} has shown that the pinball loss can fall short of achieving the quantile property on the entire dataset in the presence of regularization, suggesting the use of additional unconstrained constant terms to maintain guarantees for a discrete number of quantiles $\tau$. For subsets, work in the fairness literature \cite{Lafferty:2019} demonstrated how the quantile property can suffer over certain subsets of the population if those subsets are not known to the model, with the authors recommending learning per-group post-shifts to correct these shortcomings.
In fact, in the presence of parameter sharing, simultaneous quantile learning, and monotonicity, our proposed model structure may not be able to satisfy the quantile property everywhere, even if it has access to Boolean features defining the subsets of interest. 

We propose the use of \emph{rate constraints} to help the model achieve the quantile property for specified subsets and quantiles. Rate constraints are data-dependent constraints on metrics like accuracy or recall in an empirical risk minimization framework \cite{Goh:2016,CotterJMLR,CotterALT:2019,CotterICML:2019,PairwiseFairness}.  
We set up our rate constraints as in (\ref{eqn:fairness}) to impose that the quantile property hold over selected subsets of the training data, with some slack $\epsilon$.  The slack values are a hyperparameter of the training, and can be chosen by validation, or set by the model maker based on what they find is feasible to achieve.  This use of rate constraints may decrease the training loss, but can regularize the model to work well on the subsets of interest, and avoid overfitting noisy training examples. 

Rate constraints are non-differentiable and data dependent, and so take some care to impose: we use the open-source TensorFlow Constrained Optimization library (\ref{eqn:fairness}), using its \emph{best iterate} \cite{CotterJMLR} as an approximate solution. Note that unlike the non-crossing constraints in (\ref{eqn:peace}), which are constraints purely on the model parameters and thus can be guaranteed no matter how the model is used, rate constraints are defined on a dataset, so even if the constraints are perfectly satisfied on the training set, the rate constraints might not hold on an IID test set. Using additional validation sets (which we did not do) can improve the generalization of the constraint satisfaction \cite{CotterJMLR}.

\begin{table*}
  \caption{Simulation results: Quantile MSE and crossing violations computed for $\tau \in \{0.01, 0.02, \ldots, 0.99\}$.}
  \label{tab:sim_dist}
  \centering
  \begin{tabular}{lrrrrrrrr}
    \toprule
    & \multicolumn{2}{c}{Sine-skew (1,7)} & \multicolumn{2}{c}{Griewank} & \multicolumn{2}{c}{Michalewicz} & \multicolumn{2}{c}{Ackley} \\
    \cmidrule(lr){2-3} \cmidrule(lr){4-5} \cmidrule(lr){6-7} \cmidrule(lr){8-9}
    Model & MSE & Viol. & MSE & Viol. & MSE & Viol. & MSE & Viol. \\
    \midrule
    DLN mono & $\mathbf{3.51 \pm 0.13}$ & $0$ & $\mathbf{0.55 \pm 0.01}$ & $0$ & $\mathbf{0.219 \pm 0.006}$ & $0$ & $\mathbf{206 \pm 2}$ & $0$ \\
    DLN non-mono  & $4.41 \pm 0.09$ & $8\%$ & $0.96 \pm 0.02$ & $11\%$ & $0.265 \pm 0.006$ & $20\%$ & $455 \pm 8$ & $0$ \\
    DNN non-mono  & $\mathbf{3.53 \pm 0.09}$ & $38\%$ & $1.29 \pm 0.02$ & $5\%$ & $0.311 \pm 0.011$ & $12\%$ & $237 \pm 6$ & $0.3\%$ \\
    SQF-DNN mono  & $5.68 \pm 0.11$ & $0$ & $1.05 \pm 0.01$ & $0$ & $\mathbf{0.232 \pm 0.006}$ & $0$ & $667 \pm 81$ & $0$ \\
    QRF mono  & $6.07 \pm 0.21$ & $0$ & $1.50 \pm 0.02$ & $0$ & $0.378 \pm 0.019$ & $0$ & $343 \pm 9$ & $0$ \\
    \bottomrule
  \end{tabular}
\end{table*}

\begin{table*}
  \caption{Real data experiments: Each column is the pinball loss on the test set, averaged over $\tau \in \{0.01, 0.02, \ldots, 0.99\}$.}
  \label{tab:expt1}
  \centering
  \begin{tabular}{lrrrr}
    \toprule
    Model & Air Quality & Traffic  & Wine & Puzzles\\
    \midrule
DLN mono on $\tau$ only & $\mathbf{17.184 \pm 0.292}$ & $\mathbf{0.04778 \pm 0.00003}$ & $0.6430 \pm 0.0007$ & $3.281 \pm 0.023$\\
DLN mono on $\tau$ \& features & $\mathbf{17.184 \pm 0.292}$ & $\mathbf{0.04778 \pm 0.00003}$ & $0.6345 \pm 0.0002$ & $\mathbf{3.046 \pm 0.019}$\\
DLN non-mono& $\mathbf{17.554 \pm 0.427}$ & $\mathbf{0.04782 \pm 0.00002}$ & $0.6495 \pm 0.0008$ & $3.395 \pm 0.036$\\
DNN non-mono& $17.606 \pm 0.213$ & $0.04929 \pm 0.00030$ & $0.6421 \pm 0.0014 $ & $3.246 \pm 0.025$\\
SQF-DNN mono & $\mathbf{17.375 \pm 0.128}$ & $0.04851 \pm 0.00011$ & $0.6438 \pm 0.0027$ & $\mathbf{3.073 \pm 0.017}$\\
QRF mono & $\mathbf{17.206 \pm 0.045}$ & $0.04795 \pm 0.00001$ & $\mathbf{0.6081 \pm 0.0001}$ & $\mathbf{3.060 \pm 0.014}$ \\
    \bottomrule
  \end{tabular}
\end{table*}

\section{Experiments} \label{sec:experiments}
We first show the value of using DLNs, then use DLNs to show the value of the $P_\tau$ and rate constraints regularizers, though those contributions are function-class agnostic. Bolded table results indicate that the metric is not statistically significantly different from the best metric among the models being compared, using an unpaired t-test.

\subsection{Model and Training Details}
All hyperparameters were optimized on validation sets. We used Keras models in TensorFlow 2.2 for the unrestricted DNN comparisons that optimize (\ref{eqn:objective}) \cite{SQR:2019} as well as the spline quantile function (SQF) DNNs of \citet{Gasthaus} which optimize the same objective in a different manner while also guaranteeing non-crossing quantile estimates. For DLNs, we used the TensorFlow Lattice library \cite{TFLatticeBlogPost2020}. To train models with rate constraints, we used the custom training losses from the TensorFlow Constrained Optimization library and resolved the stochasticity of the classifier \cite{Narasimhan:2019} by taking the \emph{best iterate} \cite{TFCOBlogPost2020,CotterJMLR}.
For all DNN and DLN experiments, we use the Adam optimizer \cite{Kingma:2015} with its default learning rate of 0.001, except where noted. For DNN models, we optimized over the number of hidden layers and the hidden dimension, as well as the number of distribution keypoints for SQF-DNNs in particular. For DLN models we optimized over the number of calibration keypoints, lattice vertices, and in cases with ensembles of lattices, the number and dimensionality of base models. For both DLNs and DNNs, we additionally optimized over the number of training epochs.  Training these different models took roughly equally long. Quantile regression forests (QRF) were trained with the quantregForest R package \cite{quantregForest} and validated over the number of trees and the minimum node size. For rate constraint experiments, the slack on the constraints was also validated for the lowest quantile property violation. 

\subsection{Benchmark and Real Datasets Used}
\textbf{Air Quality:} The Beijing Multi-Site Air-Quality dataset from UCI \cite{UCI:2017}\cite{Beijing:2017} contains hourly air quality data from 12 monitoring regions around Beijing. We trained models to predict the quantiles of the PM2.5 concentration from $D=7$ features: temperature, pressure, dew point, rain, wind speed, region, and wind direction. The DLN model is an ensemble of 2-layer calibrated lattice models. We split the data by time (not IID) with earlier examples forming a training set of size 252,481, later examples a validation set of size 84,145, and most recent examples a test set of size 84,145. 

\textbf{Traffic:} This is a proprietary dataset for estimating travel time on a driving route. The DLN is a 2-layer calibrated lattice model whose inputs are 1 categorical and 3 continuous features. We used 1,000 IID examples each for training, validation, and testing, with the training examples occurring earlier in time than the validation and test examples. For all Traffic models, we optimized the Adam algorithm's learning rate and batch size.

\textbf{Wine:} We used the Wine Reviews dataset from Kaggle \cite{wine}. We predict the quantiles of quality on a 100-point scale. We used $D=42$ features, including price, country of origin, and 40 Boolean features indicating descriptive terms such as ``complex'' or ``oak''. The DLN model is an ensemble of 2-layer calibrated lattice models, each containing a subset of the features (the exact size of the ensemble and dimension of each model chosen by validation set performance). Using the DLN architecture, we further constrain the model output to be monotonically increasing in the \textit{price} feature. The data was split IID with 84,641 examples for training, 12,091 for validation, and 24,184 for testing. 

\textbf{Puzzles:} The Hoefnagel Puzzle Club uses quantile estimates of how long a member will hold a puzzle borrowed from their library before returning it. The dataset we use is publicly available on their website. Each example has five past hold-times, and a sixth feature denotes if a user belongs to one of three subsets based on their past activity, \{active users, high-variance users, new users\}. The DLN model is an ensemble of 2-layer calibrated lattice models. For the DLN, we also constrain the model output to be monotonically increasing in the most recent past hold-time feature. The 936 train and 235 validation examples are IID from past data, while the 210 test samples are the most recent samples (not IID with the train and validation data). 

\subsection{DLNs as Quantile Function Predictors}
We test our proposal to use monotonic DLNs trained with a uniform $P_{\Tau}$ expected pinball loss to predict all quantiles simultaneously. 

\textbf{Simulations:} We start with a selection of simulations from a recent quantile regression survey paper \cite{Survey:2020} based on the sine, Griewank, Michalewicz, and Ackley functions with carefully designed noise distributions to represent a range of variances and skews across the respective input domains. We used 250 training examples for the 1-D sine and Michalewicz functions, 1,000 points for the 2-D Griewank function, and 10,000 points for the 9-D Ackley function. Our metric (MSE) is the average $L^2$ difference between the estimated and true quantile curves, averaged over values of $x$ across the domain, averaged over 100 repeats. We also compute the fraction of test points for which at least two of their 99 quantiles crossed. 

The results in Table~\ref{tab:sim_dist} demonstrate that across these four disparate simulations, the proposed monotonic DLNs are the best or statistically tied for the best, and the monotonicity on $\tau$ consistently improves the performance over the non-monotonic DLN. The non-monotonic DNNs were trained with the same sampled expected pinball loss, and are sometimes close in performance but suffer substantially from crossing quantiles, despite the hypothesis in recent work that just minimizing $E_{\Tau}$ would reduce quantile crossing \cite{SQR:2019}. For example, on the sine-skew task, crossing was observed between at least two quantiles on 37.9\% of test $x$ values! Spline quantile functions \cite{Gasthaus} and quantile regression forests \cite{Meinshausen:2006} avoid crossing by construction, but performed inconsistently on the simulations.

\textbf{Real Data:} Table~\ref{tab:expt1} compares these models on the four real datasets. The DLN constrained to be monotonic on $\tau$ performed the best or statistically similar to the best on three of the four problems, and was statistically significantly better than the (non-monotonic) DNNs trained with the same expected pinball loss \cite{SQR:2019} in each case. The monotonicity constraint on the DLN slightly improved its test peformance over the unconstrained DLN for all four problems. SQF-DNNs tied the DLN with monotonicity on two of the four datasets. The QRFs proved more effective on the real datasets than in the simulations, doing particularly well on the Wine dataset, which we hypothesize is an artifact of there being only 20 possible training labels and QRFs predicting sample quantiles from subsets of the training labels.

\textbf{Feature monotonicity regularizers:} 
Table~\ref{tab:expt1} also shows that adding monotonicity constraints on relevant input features further improves the test pinball loss.  For wine, the price input was constrained to have a monotonic effect on the predicted wine quality quantiles \citep{Gupta:2018}, and for Puzzles, each of the past hold-times was constrained to have a monotonic effect on the predicted future hold time quantiles. For air quality and traffic, there were no input features that we thought should be monotonic, so those results are the same as for monotonic on $\tau$ only. Constraining input features to be monotonic also helps explain what the model is doing \cite{GuptaEtAl:2016}. 

\subsection{Training with a Smoothed Expected Pinball Loss}
We test our proposal that minimizing an expected pinball loss with $P_{\Tau}$ provides useful regularization.

\textbf{Unconditioned:} We start with the classic problem of predicting the quantiles an exponential distribution with $\lambda = 1$ without any features $X$ to condition on. We set the DLN to simply be a linear model on $\tau \in [0,1]$, trained to minimize the expected pinball loss over a Beta $P_{\Tau}$ with mode set at the desired quantile. We compare to the sample quantile, which minimizes the pinball loss for its $\tau$, and to the Harrell-Davis estimator \cite{HarrellDavis:1982}. 

\begin{figure}
\centering
\begin{subfigure}{0.9\linewidth}
\centering
\includegraphics[width=\linewidth]{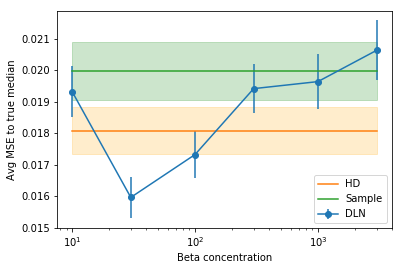}
\subcaption{$N=51,\ \tau=0.5$}
\label{fig:exp_q50}
\end{subfigure}
\begin{subfigure}{0.9\linewidth}
\centering
\includegraphics[width=\linewidth]{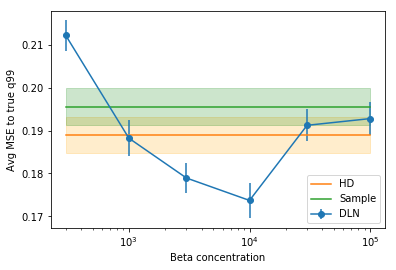}
\subcaption{$N=505,\ \tau=0.99$}
\label{fig:exp_q99}
\end{subfigure}
\caption{\textbf{Unconditional Quantile Estimation of $\exp(\lambda = 1)$} Sample is the sample quantile, HD is the Harrell-Davis estimator, and DLN uses a a linear model $f(\tau)$ on $\tau \in [0,1]$ fit to minimize expected pinball loss with respect to Beta $P_{\Tau}$.  Results were averaged over 1,000 random draws of $N$ samples, with 95\% confidence intervals depicted by shading and error bars.}
\label{fig:exp_fig}
\end{figure}

Figure~\ref{fig:exp_fig} shows that for high Beta concentrations (producing a spiky $P_{\Tau}$ on the quantile of interest), the DLN performs similarly to the sample quantile, as expected. For a middle range of Beta concentrations, minimizing the expected pinball loss with $P_{\Tau}$ beats the Harrell-Davis estimator.  

\begin{figure*}[p]
\centering
\begin{subfigure}{0.32\textwidth}
  \centering
  \includegraphics[width=\linewidth]{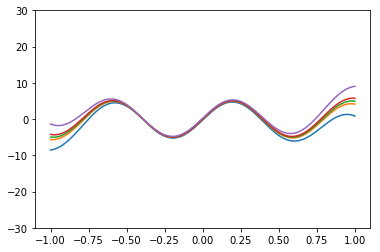}
  \subcaption{Sine-skew (1,1)}
\end{subfigure}
\begin{subfigure}{0.32\textwidth}
  \centering
  \includegraphics[width=\linewidth]{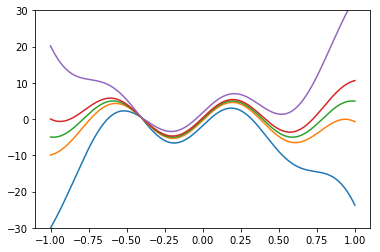}
  \subcaption{Sine-skew (7,7)}
\end{subfigure}
\begin{subfigure}{0.32\textwidth}
  \centering
  \includegraphics[width=\linewidth]{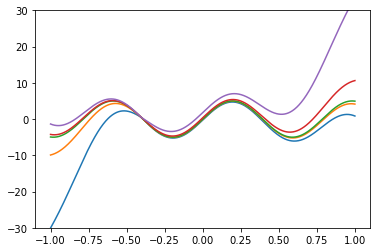}
  \subcaption{Sine-skew (1,7)}
\end{subfigure}
\caption{True quantiles of the sine-skew distribution with different noise parameters \cite{Survey:2020}. The colored lines show the quantiles for $\tau = 0.1, 0.4, 0.5, 0.6, 0.9$.}
\label{fig:sinskew}
\end{figure*}

\textbf{Single Feature:} We ran simulations on the 1D sine-skew function of \citet{Survey:2020} plotted in Figure~\ref{fig:sinskew} with noise parameters $(a,b) = (1, 1)$ for symmetric low-noise, $(7, 7)$ for symmetric high-noise, and $(1, 7)$ a sharply asymmetric high-noise function.

Table~\ref{tab:sinskew} shows that in the $(1, 1)$ low-noise case the high-smoothing options with the Beta $P_{\Tau}$ concentration hyperparameter at $C=2$ (uniform) and $C=10$ work best. In the $(7, 7)$ case of high heteroskedasticity, we are best off with moderate amounts of smoothing: the quantiles nearby to $\tau=0.5$ resemble the well-behaved $(1, 1)$ case. In the $(1, 7)$ case, the erratic asymmetric tails resist smoothing; they introduce enough model error that we are best off essentially training for the median alone. In general, more training data improves the relative performance of low-smoothing models.

\begin{table*}[p]
\caption{Sine-skew experiment with different sine-skew noise choices (1,1), (7,7) and (1,7). MSE between true median and estimated median, averaged over $x \sim \textrm{Unif}(-1,1)$. Trained with expected pinball loss with Beta $P_{\Tau}$ with mode $\tau = 0.5$ and varying concentrations $C$.  Note that $C=2$ is the uniform distribution while $C=10{,}000$ is close to single-$\tau$ sampling.}
  \label{tab:sinskew}
  \centering
  \begin{tabular}{llccccc}
  \toprule
  $(a,b)$ & $N$ & $C = 2$ & $C = 10$ & $C = 100$ & $C = 1{,}000$ & $C = 10{,}000$ \\
  \midrule
  $(1,1)$ & $100$ & $\mathbf{0.456 \pm 0.032}$ & $\mathbf{0.446 \pm 0.030}$ & $\mathbf{0.509 \pm 0.038}$ & $\mathbf{0.495 \pm 0.025}$ & $0.633 \pm 0.047$ \\
  $(1,1)$ & $1{,}000$ & $\mathbf{0.070 \pm 0.003}$ & $\mathbf{0.073 \pm 0.004}$ & $\mathbf{0.075 \pm 0.004}$ & $\mathbf{0.073 \pm 0.003}$ & $0.078 \pm 0.004$ \\
  \midrule
  $(7,7)$ & $100$ & $\mathbf{6.973 \pm 0.289}$ & $\mathbf{6.193 \pm 0.345}$ & $7.534 \pm 0.368$ & $\mathbf{7.114 \pm 0.349}$ & $7.568 \pm 0.363$ \\
  $(7,7)$ & $1{,}000$ & $4.330 \pm 0.148$ & $1.528 \pm 0.077$ & $\mathbf{1.164 \pm 0.048}$ & $\mathbf{1.111 \pm 0.051}$ & $1.281 \pm 0.068$ \\
  \midrule
  $(1,7)$ & $100$ & $3.796 \pm 0.248$ & $\mathbf{2.918 \pm 0.180}$ & $\mathbf{2.942 \pm 0.174}$ & $3.268 \pm 0.223$ & $\mathbf{2.557 \pm 0.178}$ \\
  $(1,7)$ & $1{,}000$ & $0.950 \pm 0.059$ & $0.588 \pm 0.036$ & $\mathbf{0.480 \pm 0.025}$ & $\mathbf{0.471 \pm 0.026}$ & $\mathbf{0.415 \pm 0.032}$ \\
  \bottomrule
  \end{tabular}
\end{table*}

\begin{table*}[p]
  \caption{Effect of training with different sampling distributions for $\tau \sim P_{\Tau}$. The key comparisons are: (1) the single models trained with Unif vs Discrete, and (2) the three separate models trained with Beta or single $\tau$. All models were DLNs and monotonic on $\tau$.}
  \label{tab:expt2All}
  \centering
\begin{tabular}{rlrrr}
  \toprule
  \textbf{Air Quality:} & Model   &  Pinball loss ($\tau = 0.5$) & Pinball loss ($\tau = 0.9$) & Pinball loss ($\tau = 0.99$) \\
  \cmidrule{2-5}
  & $\tau \sim \textrm{Unif}(0,1)$ & $\mathbf{23.576 \pm 0.047}$ & $14.850 \pm 0.075$ & $2.947 \pm 0.025$ \\
  & $\tau \sim \textrm{Discrete}$ & $24.083 \pm 0.073$ & $15.439 \pm 0.062$ & $3.042 \pm 0.012$  \\
  & $\tau \sim \textrm{Beta}$ & $\mathbf{23.586 \pm 0.183}$ & $\mathbf{13.942 \pm 0.054}$ & $\mathbf{2.709 \pm 0.029}$  \\
  & Single $\tau \in \mathbf{T}$ & $\mathbf{23.634 \pm 0.068}$ & $14.908 \pm 0.092$ & $\mathbf{2.700 \pm 0.013}$ \vspace{1.0 ex} \\
  \textbf{Traffic:} & Model & Pinball loss ($\tau = 0.5$) & Pinball loss ($\tau = 0.9$) & Pinball loss ($\tau = 0.99$) \\
  \cmidrule{2-5}
  & $\tau \sim \textrm{Unif}(0,1)$ & $\mathbf{0.064386 \pm 0.00014}$ & $0.040159 \pm 0.00018$ & $\mathbf{0.010645 \pm 0.00018}$ \\
  & $\tau \sim \textrm{Discrete}$ & $\mathbf{0.064578 \pm 0.00028}$ & $\mathbf{0.039804 \pm 0.00014}$ & $0.011290 \pm 0.00014$ \\
  & $\tau \sim \textrm{Beta}$ & $\mathbf{0.064988 \pm 0.00030}$ & $\mathbf{0.039549 \pm 0.00009}$ & $\mathbf{0.01064 \pm 0.00011}$ \\
  & Single $\tau \in \mathbf{T}$ & $\mathbf{0.064791 \pm 0.00012}$ & $0.039900 \pm 0.00010$ & $\mathbf{0.01070 \pm 0.00018}$ \vspace{1.0 ex} \\
    \textbf{Wine:} & Model & Pinball loss ($\tau = 0.1$) & Pinball loss ($\tau = 0.5$) & Pinball loss ($\tau = 0.9$) \\
    \cmidrule{2-5}
    & $\tau \sim \textrm{Unif}(0,1)$ & $0.4099 \pm 0.0006$ & $\mathbf{0.8889 \pm 0.0017}$ & $0.3773 \pm 0.0019$  \\
    & $\tau \sim \textrm{Discrete}$ & $0.4094 \pm 0.0005$ & $0.8891 \pm 0.0010$ & $0.3706 \pm 0.0003$  \\
    & $\tau \sim \textrm{Beta}$ & $\mathbf{0.4047 \pm 0.0005}$ & $\mathbf{0.8871 \pm 0.0008}$ & $\mathbf{0.3665 \pm 0.0004}$  \\
    & Single $\tau \in \mathbf{T}$ & $\mathbf{0.4049 \pm 0.0004}$ & $\mathbf{0.8867 \pm 0.0006}$ & $\mathbf{0.3663 \pm 0.0003}$    \vspace{1.0 ex} \\
    \textbf{Puzzles:} & Model   &  Pinball loss ($\tau = 0.5$) & Pinball loss ($\tau = 0.7$) & Pinball loss ($\tau = 0.9$) \\
    \cmidrule{2-5}
    & $\tau \sim \textrm{Unif}(0,1)$ & $\mathbf{4.173 \pm 0.013}$ & $\mathbf{4.219 \pm 0.021}$ & $2.705 \pm 0.029$ \\
    & $\tau \sim \textrm{Discrete}$ & $4.359 \pm 0.014 $ & $\mathbf{4.204 \pm 0.011}$ & $\mathbf{2.614 \pm 0.010}$  \\
    & $\tau \sim \textrm{Beta}$ & $4.318\pm 0.013$ & $4.277 \pm 0.018$ & $2.681 \pm 0.011$  \\
    & Single $\tau \in \mathbf{T}$ & $4.293 \pm 0.013$ & $4.350 \pm 0.022$ & $2.708 \pm 0.012$ \\
    \bottomrule
  \end{tabular}
\end{table*}

\textbf{Real Data:} Table~\ref{tab:expt2All} compares the accuracy of using different distributions $P_{\Tau}$ to fit three target quantiles on each dataset. First, we compare two methods of training a single model to fit all three quantiles: $\tau \sim \textrm{Unif}(0,1)$, as in \citet{SQR:2019}, vs $\tau \sim \textrm{Discrete}$ on just the three target quantiles. These are competitive with each other, on average.

Next, we compare training three separate models to fit the three quantiles either (1) using a Beta distribution centered on the target quantile and a concentration hyperparameter validated within [10, 1000], or (2) using only the single target quantile itself. The Beta-smoothed models performed as well or better than the single-quantile models in all cases.

At the extreme, training for single $\tau$s using uniform $P_{\Tau}$ works well, sometimes better, sometimes worse than single, and has the added advantage of providing a complete inverse CDF in a single model without quantile-crossing.

\subsection{Effect of Rate Constraints}
We use rate constraints to ensure the quantile property roughly holds across subsets of interest on the training data. For the Air Quality and Traffic problems, we apply constraints on $\tau \in \mathbf{T} = \{0.5,0.9,0.99\}$, emphasizing accurate estimates of upper quantiles for air pollution and traffic for the 12 regions in the Air Quality dataset and the 10 countries in the Traffic dataset.  For the Wine problems, we apply constraints over $\tau \in \mathbf{T} = \{0.1,0.5,0.9\}$ for 20 countries plus an ``Other country'' category aggregating the remaining small countries, highlighting both the upper and lower quantiles of wine quality. For the Puzzles problem, we apply constraints over $\tau \in \mathbf{T} = \{0.5,0.7,0.9\}$, and enforce constraints over three subsets of users: \{active users, high-risk users, new users\}.

For each problem, we included a rate constraint for each combination of the subsets of interest and the quantiles of interest (for example, for 10 countries and 3 quantiles of interest, there are $10 \times 3 = 30$ rate constraints). Hyperparameters were chosen from the validation performance according to the heuristic from \citet{CotterJMLR} that considers both constraint violation and objective. We compare against unconstrained models that are trained and validated to optimize pinball loss. The max quantile violation metric takes the max of the absolute quantile error over all constrained subsets $\mathcal{D}_s$ and quantiles $\tau_s$:
\begin{align*}
\displaystyle \max_{s \in \{1, \ldots, S\}}\left|\tau_s - \frac{1}{|\mathcal{D}_s|} \sum_{(x_j, y_j) \in \mathcal{D}_s} \mathbb{I}[y_j \leq f(x_j, \tau_s; \theta)]\right|.
\end{align*}

Table~\ref{tab:expt3All} reports the max quantile violation over the subsets on the quantiles $\mathbf{T}$ on the test set, and pinball loss averaged over the quantiles $\mathbf{T}$ on the test set. The rate constraints significantly improved the test maximum quantile violation for Air Quality, Wine, and Puzzles, and was statistically tied for Traffic. These subset wins caused mixed results on the overall test pinball loss: statistically significantly hurting it for Air Quality, but statistically significantly improving the overall pinball loss on Puzzles, which is the most non-IID of the four datasets, with the test set known to have more examples from the hard \emph{new-users} subset which may have benefited from its rate constraint at training time. 

\begin{table}[ht!]
 \caption{Effect of rate constraints. \textbf{Unconstr} is a DLN with no rate constraints. \textbf{Constr} is a DLN with rate constraints.}
 \label{tab:expt3All}
 \centering
\begin{tabular}{lrr}
    \toprule
& \multicolumn{2}{c}{Air Quality}\\ 
\cmidrule{2-3}
 Model & Max violation & Pinball loss \\
    \midrule
    Unconstr  & $0.186 \pm 0.002$  & $\mathbf{15.112 \pm 0.034}$ \\
    Constr & $\mathbf{0.170 \pm 0.004}$ & $15.336 \pm 0.013$ \vspace{1.0 ex} \\
& \multicolumn{2}{c}{Traffic} \\
\cmidrule{2-3}
Model & Max violation & Pinball loss \\
 \midrule
 Unconstr & $\mathbf{0.0929 \pm 0.0035}$ & $\mathbf{0.0382 \pm 0.00007}$ \\
 Constr & $\mathbf{0.0917 \pm 0.0022}$ & $\mathbf{0.0382 \pm 0.00006}$ \vspace{1.0 ex} \\
& \multicolumn{2}{c}{Wine} \\
\cmidrule{2-3}
Model & Max violation & Pinball loss \\
    \midrule
    Unconstr & $0.2077 \pm 0.0192$ & $\mathbf{0.6401 \pm 0.0062}$ \\
    Constr & $\mathbf{0.1774 \pm 0.0069}$ & $\mathbf{0.6393 \pm 0.0008}$ \vspace{1.0ex} \\
& \multicolumn{2}{c}{Puzzles} \\
\cmidrule{2-3}
Model & Max violation & Pinball loss \\
    \midrule
    Unconstr & $0.347 \pm 0.005$ & $3.701 \pm 0.012$ \\
    Constr & $\mathbf{0.326 \pm 0.008}$ & $\mathbf{3.588 \pm 0.013}$ \\
    \bottomrule
 \end{tabular}
\end{table}

\section{Conclusions} \label{sec:conclusions}
We investigated different regularization strategies for quantile regression. First, we built on classic statistics about sample quantile estimation to propose training with a smoothed expected pinball loss. We showed that a uniform $P_{\Tau}$ yields performance similar to that of discrete quantiles, plus a more flexible model that can predict any $\tau$. We demonstrated that smoothing with a Beta $P_{\Tau}$ can be more accurate than training for a single $\tau$ of interest. 

We then attacked the classic goal of non-crossing quantiles with DLNs and showed that DLNs with $\tau$ and feature monotonicity work well for quantile regression, and do so in an empirical risk minimization (ERM) framework, with all the flexibility and computational efficiency the ERM framework brings. Not only do they provide a way to guarantee non-crossing for multi-quantile regression without limiting flexibility, they also proved effective at predicting the full conditional distribution of $y$ given $x$ across a wide variety of problems.

Lastly, we showed that rate constraints on subsets of data can improve test performance on those subsets, and may help or hurt the aggregate loss.  Here, we used rate constraints to buoy the worst-case of the subsets, a common fairness notion. As these strategies attack different aspects of the problem, they can be used separately or work together.

\clearpage

\bibliographystyle{icml2021}
\bibliography{references}

\clearpage

\appendix

\section{Background on Deep Lattice Networks}
We provide more background on Deep Lattice Networks (DLNs) \cite{You:2017}. DLNs are an arbitrarily flexible function class that can efficiently impose monotonicity on any subset of input features, without restricting the flexibility on other features. They are constructed by composing together layers that individually preserve monotonicity, the most notable of which is the lattice layer.

A lattice \cite{Garcia:09} is a function parameterized by the values $\theta$ it takes at knots arrayed in a regular grid throughout the input domain, which is assumed to be bounded. The function value for all other inputs is generated by linearly interpolating from the surrounding knots. 

The simplest example of a lattice is a 1-dimensional lattice, equivalent to a piecewise-linear function (PLF). PLFs play an important role in DLNs; they are often used to individually transform inputs before they are fed into more complex lattices.

Given a $D$-dimensional input with $L_d$ knots for the $d$th input, we therefore have $\theta \in \mathbb{R}^L$, where $L = \prod_{d=1}^D L_d$. Given an interpolation kernel $\Phi: \mathbb{R}^D \rightarrow \mathbb{R}^L$, the lattice function $f$ can be expressed as a kernel function:

$$f(x) = \sum_{i=1}^L \Phi(x)_i \theta_i$$

Our interpolation strategy is the multilinear method discussed in \citet{GuptaEtAl:2016}. At a high level, this means that the weight on a knot is the product of the scalar weights we'd place on that knot in each dimension. In particular, let $v_d$ be the vector of knot positions in the $d$th dimension and $v_d[l]$ and $v_d[r]$ be the knots on either side of $x_d$. Our right-weight $w_d[r]$ would be $\frac{x_d - v_d[l]}{v_d[r] - v_d[l]}$ and our left-weight $w_d[l] = 1 - w_d[r]$. Our final interpolation weight on a particular knot $\theta_i$ is either 0 (if it is not in the surrounding hypercube) or $\prod_{d=1}^D w_d[s]$, where $s$ corresponds to the left- or right-weight depending on whether the knot is to the left- or right- of the input $x$ in that dimension.

\begin{figure}
\centering
\includegraphics[width=1.0\linewidth]{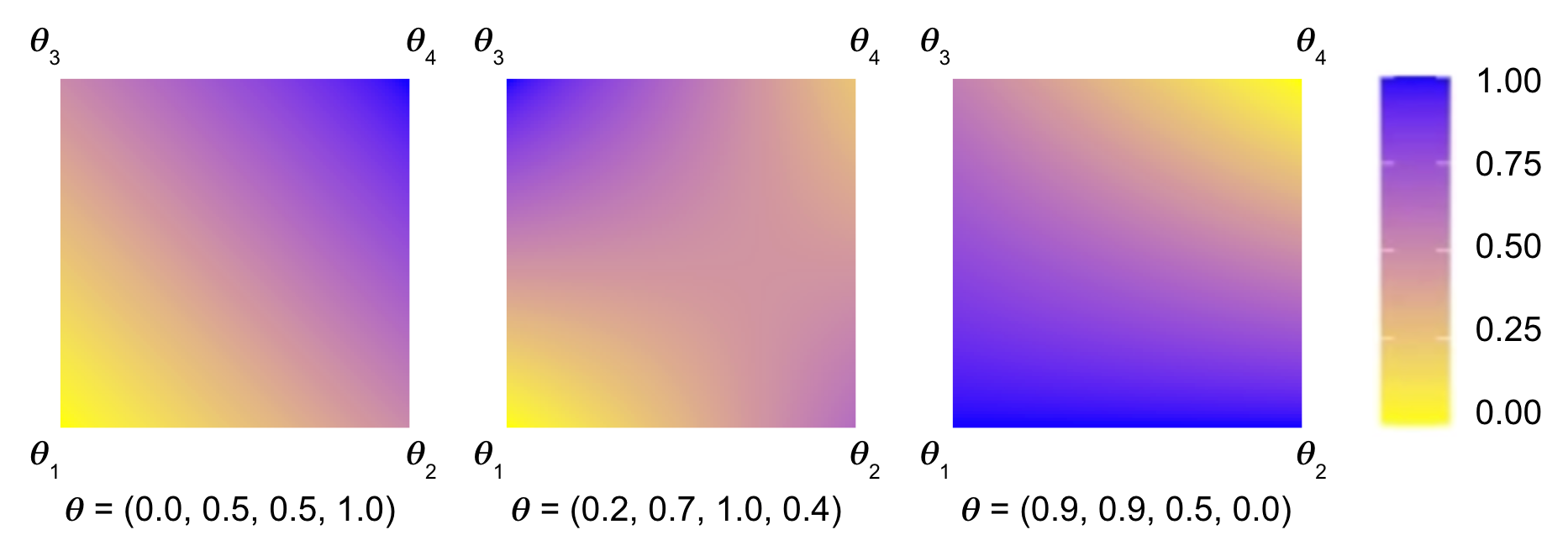}
\caption{Example 2-d lattices with four parameters, each of which represents the value the function takes at one of the four corners of the input domain. Intermediate values are computed via multilinear interpolation.}
\label{fig:sample-lattices}
\end{figure}

We show some examples of 2-dimensional lattices with $2 \times 2$ knots for a total of four parameters in Figure~\ref{fig:sample-lattices}. You can see how lattices are capable of learning interactions between features, and that given sufficient knots over the bounded input domain, lattice models can approximate any continuous bounded input-output relationship.

A few key ideas make lattices generally useful and easy to use. The first is that they are differentiable in their parameters and so can be learned with any standard gradient-based approach in an empirical risk minimization framework. 

The second is that their parameterization makes it straightforward to impose constraints such as monotonicity by imposing that any two neighboring parameters in the selected direction in the look-up table obey the monotonicity constraints. In Figure~\ref{fig:sample-lattices}, that would entail constraining $\theta_2 \ge \theta_1$ and $\theta_4 \ge \theta_3$ to achieve monotonicity in the horizontal dimension.  We then solve the overall machine learning task as a constrained optimization problem with, for example, a Projected Stochastic Gradient Descent algorithm.

Finally, it is straightforward to create ensembles of lattices, which would otherwise be constrained by their exponential number of parameters in the number of features.

We use the open-source TensorFlow Lattice package \cite{TFLatticeBlogPost2020} in our experiments.

\end{document}